\documentclass[a4paper, 11pt]{article}
\usepackage[utf8]{inputenc}
\usepackage[margin=2cm]{geometry} 

\usepackage{lineno}

\usepackage{geometry}
\usepackage{amsmath}
\usepackage{amsfonts}
\usepackage{graphicx, multirow}
\usepackage{makecell}
\usepackage{siunitx}
\usepackage{stmaryrd}

\usepackage{subcaption}
\usepackage{arydshln}
\usepackage[hidelinks]{hyperref}

\usepackage{newfloat}
\DeclareFloatingEnvironment[name={Supplementary Figure}]{suppfigure}
\DeclareFloatingEnvironment[name={Supplementary Table}]{supptable}

\usepackage[tableposition=top, font=footnotesize, labelfont=bf]{caption}

\usepackage{pdfpages}
\usepackage{verbatim}

\makeatletter
\def\blfootnote{\xdef\@thefnmark{}\@footnotetext}
\makeatother
\newcommand{\ie}{\emph{i.e.\ }}
\newcommand{\eg}{\emph{e.g.\ }}

\begin{document}
\begin{center}
    \Large \textbf{Probabilistic Spatial Analysis in Quantitative Microscopy\\ with Uncertainty-Aware Cell Detection using\\ Deep Bayesian Regression of Density Maps}\\\Large 
    
    \vspace{.5em}
    \small Alvaro Gomariz$^{1,2,*}$, Tiziano Portenier$^1$, C\'esar Nombela-Arrieta$^{2}$, Orcun Goksel$^{1,3}$ \\ 
    \vspace{.5em}
    \footnotesize
    $^1$ Computer-assisted Applications in Medicine, Computer Vision Lab, ETH Zurich, Switzerland\\
    $^2$ Department of Medical Oncology and Hematology, University Hospital and University of Zurich, Switzerland\\
    $^3$ Department of Information Technology, Uppsala University, Sweden
    \blfootnote{$^*$ Corresponding author e-mail: alvaroeg@ethz.ch}
\end{center}

\begin{abstract}
3D microscopy is key in the investigation of diverse biological systems, and the ever increasing availability of large datasets demands automatic cell identification methods that not only are accurate, but also can imply the uncertainty in their predictions to inform about potential errors and hence confidence in conclusions using them. 
While conventional deep learning methods often yield deterministic results, advances in deep Bayesian learning allow for accurate predictions with a probabilistic interpretation in numerous image classification and segmentation tasks. 
It is however nontrivial to extend such Bayesian methods to cell detection, which requires specialized learning frameworks.
In particular, regression of density maps is a popular successful approach for extracting cell coordinates from local peaks in a postprocessing step, which then, however, hinders any meaningful probabilistic output. 
We herein propose a deep-learning based cell detection framework that can operate on large microscopy images and outputs desired probabilistic predictions by
($i$)~integrating Bayesian techniques for the regression of uncertainty-aware density maps, where peak detection algorithms can be applied to generate cell proposals, and
($ii$)~learning a mapping from the numerous proposals to a probabilistic space that is calibrated, \ie accurately represents the chances of a successful prediction. 
Utilizing such calibrated predictions, we propose a probabilistic spatial analysis with Monte-Carlo sampling.
We demonstrate this in revising an existing quantitative description of the distribution of a mesenchymal stromal cell type within the bone marrow, where our proposed methods allow us to reveal spatial patterns that are otherwise undetectable. 
Introducing such probabilistic analysis in quantitative microscopy pipelines will allow for reporting confidence intervals for testing biological hypotheses of spatial distributions.
\end{abstract}

\section*{Introduction}
Advances in microscopy imaging currently enable the inspection of cells within biological tissues with astonishing resolution.
In particular, fluorescence microscopy (FM) offers the possibility to separately examine distinct cellular structures stained with carefully selected markers.
Visual inspection of such FM datasets in the context of large 3D images of intact tissues has revealed new information on the structures and mechanisms of different biological tissues~\cite{gomariz2019imaging}.
Effective quantitative description of cellular distributions in such large datasets relies on the accuracy of cell detection, which is the task of identifying the presence and location of cells in the images. 
Despite the advances in bioimage analysis suites~\cite{Schindelin2012, Jones2008, Sommer2011}, this task remains challenging~\cite{Meijering2016, Sbalzarini_2016} and is often addressed using burdensome and time-consuming manual annotations~\cite{gomariz2019imaging}.

Cell detection has been a fundamental challenge in the spatial characterization of hematopoiesis, the process by which blood cells are created, which mostly occurs within bone marrow (BM) tissues~\cite{Nombela_Arrieta_2017}. 
Hematopoiesis takes place within a structural framework or microenvironment provided by non-hematopoietic, so-called stromal cells, which critically modulate the behaviour of hematopoietic progenitors through specific interactions in restricted anatomical spaces denominated \emph{niches}~\cite{Mendelson_2014, Scadden_2012, Calvi_2013}. A constantly increasing number of stromal cell subsets has been described to date and their accurate identification and detection within microscopy images is key in the investigation of their functional features and dynamics in health and disease.
As for all complex multicellular tissues, attempts for understanding spatial patterns and cellular interactions that underlie organ function in the bone marrow rely critically on the quantification of distances between cellular coordinates as well as to larger supracellular anatomical structures of interest, such as blood vessels or bone surfaces~\cite{gomariz2019imaging,Kiel_2005, Nombela_Arrieta_2013, Zhao2014}.

A framework for quantification of spatial distributions was proposed in~\cite{gomariz2018quantitative}, with the use of point processes~\cite{Baddeley_2015, Helmuth_2010, Lagache_2013, Jammalamadaka_2015}.
This was demonstrated for the spatial analysis of two key stromal components within bone marrow sections: the \emph{sinusoids} that form the microvasculature of the BM, and a pool of mesenchymal cells of fibroblastic morphology termed CXCL12-abundant reticular cells (\emph{CAR cells}).
The proposed spatial analysis was used to confirm the presumed preferential localization of CAR cells near sinusoids, for which no quantitative evidence had existed thus far. 
A number of descriptive statistics were reported in \cite{gomariz2018quantitative} as reference spatial descriptors of this biological system. 
However, although the analysis in that report provided well-established statistical methods for hypothesis testing, its strength was limited in that only the variability of the results across samples was taken into account, but the analysis did not reflect any potential mistakes in the cell detection method employed. In addition, the simplistic image processing techniques employed were time-consuming and inherently limited in accuracy.

Automatic cell detection is widely studied topic.
A group of methods utilize instance segmentation algorithms (e.g.\ watershed~\cite{Hodneland2013, WAHLBY2004, Malpica1997, Yan2008}) that assign each pixel of the image a label (cell type or background) and an identifier unique to each of the cell instances. 
Although this can allow for describing cell morphology, most spatial analyses only require localization and hence center coordinates alone, thereby rendering \emph{detection} algorithms more suitable for this purpose. 
While conventional blob detection methods~\cite{Lindeberg1998, Peng2009} are still employed in biological studies due to their simplicity~\cite{Sbalzarini_2016, Takaku2010, acar2015deep, Myers2013}, supervised deep learning (DL) approaches are slowly establishing themselves as the state of the art. 
The advantage of detection over instance segmentation becomes even more prominent for DL methods, where the effectiveness is determined primarily by the availability of high-quality manual annotations, since cell coordinate annotations requires far less effort than dense annotation of individual pixels contained within cells.

Unlike other discriminative tasks, \eg segmentation or classification, where end-to-end DL approaches have been widely adopted~\cite{falk2019, He2016}, object detection implies an underlying structure among output elements, making it a set prediction problem, for which the application of typical convolutional neural networks (CNNs) solutions is hampered~\cite{Stewart2016}.  
As a result, several alternatives have recently been proposed~\cite{Girshick2014, Uijlings2013, He2015, Girshick2015, Ren2017, Redmon2016, Liu2016a,YOLOv2_Redmon_2017, redmon2018yolov3}, where CNNs are employed for extracting informative features, which are then subsequently processed to infer object coordinates or bounding boxes. 
Although these approaches have achieved remarkable results in natural images such as ImageNet~\cite{deng2009imagenet}, they are highly dependent on post-processing steps that hinder their implementation and usability~\cite{Zhao2019a}.

If the objects to be detected are of relatively similar size and/or their exact extents are irrelevant for the subsequent analysis to be conducted (\eg cell counting or distances to each other or to surrounding structures), then bounding boxes are not required.
In these cases, the shortcomings described above can be circumvented by using CNNs to regress density maps (DMs) generated from ground truth (GT) annotations of cell locations~\cite{Xie2018}.
Cell coordinates are then detected within these DMs as peaks, for instance with non-maximum suppression (NMS) algorithms. 
Such a framework then enables the adoption of typical CNN architectures commonly employed in segmentation (\eg UNet~\cite{falk2019}) for the detection of cells and other landmarks, as utilized in recent works~\cite{He2021, Sierra2020, Xie2016, Liu2019, Zheng2020, Gomariz2019, Youn2020}.
Nevertheless, DM regression with CNNs poses two major caveats: ($i$)~detection of local peaks introduces additional hyperparameters that must be tuned for different applications, and ($ii$)~detection outputs have no probabilistic interpretation, hence uncertainties in predictions cannot be known. 

\emph{Calibration} is the term that refers to how well a probabilistic output value indicates actual probability of prediction success.
For instance, a model that identifies 100 cells all with a probability of 0.6 is said to be perfectly calibrated if 60 of these cells turn out to be correct according to some GT, and the other 40 are incorrect.
Although typical segmentation CNNs already output values that can be interpreted as prediction probabilities, most DL approaches are known to be poorly calibrated, invalidating any such probabilistic interpretation~\cite{Guo2017}.
Bayesian approaches that have traditionally excelled at confidence calibration have been recently incorporated in common CNN architectures.
These deep Bayesian learning methods have been shown to result in better calibration by accounting for two different types of uncertainties~\cite{kendall2017uncertainties}.
\emph{Epistemic} uncertainty is the lack of confidence of a model on its parameters and it may be reduced by leveraging additional labeled data.
For estimating epistemic uncertainty, a technique called Monte Carlo (MC) Dropout has been demonstrated to be a good Bayesian approximator of Gaussian process models~\cite{gal2015dropout,gal2015cnn}, and has later been incorporated into segmentation CNNs to take pixel-wise uncertainties into account while improving the quality of their predictions~\cite{bayesianSegNet17}.
This technique was inspired by dropout layers that were originally proposed as a regularization method to mitigate overfitting by randomly removing network parameters during training~\cite{srivastava2014dropout}.
MC dropout instead employs random removal after training during the \emph{inference} phase, which enables \emph{epistemic} uncertainty estimation.
\emph{Aleatoric} uncertainty is implicit to the data and cannot be avoided.
Its estimation was proposed in~\cite{kendall2017uncertainties} by utilizing a loss function that accounts for noise associated with observations while  simultaneously predicting this as well.

Bayesian DL models on medical images have been employed to improve calibration of segmentation confidence~\cite{Mehrtash2020} and to predict uncertainties as a proxy to estimate segmentation quality on previously unseen data~\cite{epistemicprobs_hiasa2019automated}.
The advantages of simultaneously accounting for both \emph{epistemic} and \emph{aleatoric} uncertainties have been studied recently in FM images~\cite{gomariz2021uncertaintymarkers}, indicating that uncertainty estimation is not only beneficial for superior segmentation results but also for an accurate estimation of prediction quality on previously unseen data. 

In this work, we build on the aforementioned fields by designing a probabilistic DL framework for calibrated DM-based cell detection, demonstrated in the context of large 3D FM images.
We validate our methods on the detection of CAR cells included in the BM dataset described in~\cite{gomariz2018quantitative}.
In earlier work, the complex 3D morphology of BM and varying intensities of FM had hindered an automatic and accurate detection of such cells, which we overcome with the methods presented herein, while furthermore demonstrating probabilistic techniques for hypothesis testing.

Below we first give a short overview of conventional CNN-based cell detection using DMs, which we later improve with a comparative analysis for different tiling strategies and design choices.
Next, we present a novel strategy with the integration of deep Bayesian methods for probabilistic classification of cell proposals, which leads to a substantial improvement in the calibration of detection results.
We finally apply our cell detection method to an extended BM stroma dataset for the spatial characterization of CAR cells, revising our previously reported results.
This study highlights the benefits of employing well calibrated models allowing for a probabilistic analysis that accounts for the confidence associated to each predicted cell.
This leads to findings that would be overlooked by conventional deterministic analysis.

\section*{Results}

\subsection*{Cell detection in density maps}
\begin{figure}[t]
    \centering
    \includegraphics[width=1\textwidth]{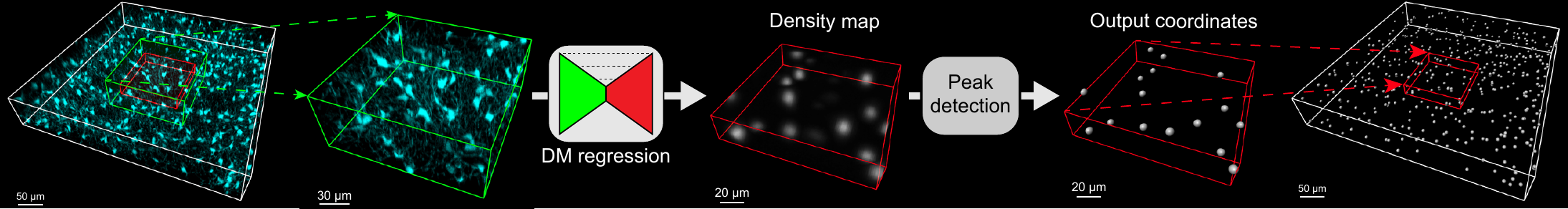}
    \caption{Illustration of our pipeline for cell detection.
    Large 3D FM samples are decomposed in patches (green frame) subsequently processed by a CNN that regresses an output DM. 
    A peak detection algorithm is applied on the DM to obtain the locations of cells.
    The resulting coordinates are then reconstructed within the original large sample volume based on the limits of the output patch (red frame) where they are contained. 
    }
    \label{fig:pipeline}
\end{figure}

We study the problem of cell detection on a 3D FM dataset of BM stroma samples, as illustrated in our overview Fig.\ \ref{fig:pipeline}.
These samples are decomposed into different patches of images acquired from CXCL12-GFP, a fluorescent protein employed for the visualization of CAR cells (details in \textit{Methods - \nameref{sec:dataset}}). 
In order to quantify the quality of detection methods, an assignment is needed between predicted cell coordinates and GT annotations, such that they can be designated as true positives (TP), false positives (FP), or false negatives (FN).
Simple assignment strategies such as nearest neighbours may lead to bias in method evaluations, \eg by assigning multiple detections to a single GT and vice versa -- a common but undesirable scenario~\cite{Stewart2016}. 
We herein employ the Hungarian algorithm, which finds optimal one-to-one assignments based on all relative distances between predictions and GT, as illustrated in Supplementary Fig.\ \ref{suppfig:hungarian_matching}.
The resulting metrics are reported following a 4-fold cross-validation approach (more details in \textit{Methods - \nameref{sec:training_evaluation}}).

Since DL-based regression of DMs has previously been successful in the detection of cell-like objects, we adopt this framework to localize cells as peaks in predicted DMs, whose values are related to the likelihood of occurrences.
However, we need to prevent detection of multiple coordinates within a neighbourhood where the typical spatial extent of a cell would physically prevent the presence of another. 
To this end, we employ NMS to iteratively detect cells (peaks) while avoiding those closer than average cell radius (details in Section \textit{Methods - \nameref{sec:dmdesign_methods}}).
Not to detect small prediction noise as local peaks, iterative NMS process is typically terminated with a stopping criterion when no peak is left above a DM threshold, which is an empirically set hyperparameter often without further analysis in the literature~\cite{Xie2018}, thus we evaluate it for different DM design strategies in the next section.

\subsection*{Tiling strategy and density map design for large images}
\label{sec:dmdesign}
\begin{figure}[p]
    \centering
    \includegraphics[width=\textwidth]{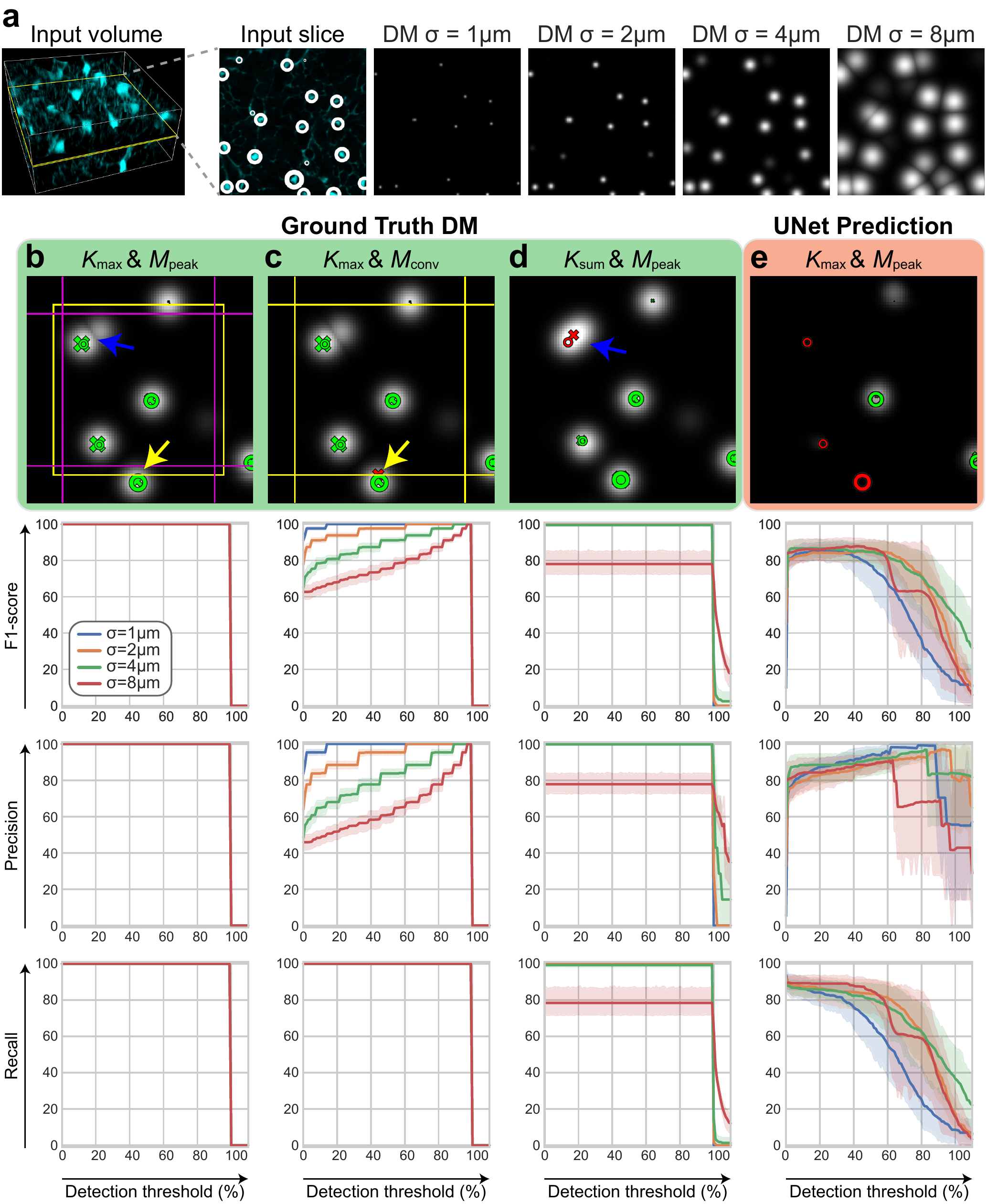}
    \caption{Analysis of DM design alternatives for regression-based cell detection. 
    (\textbf{a}) Example slice of DMs created with Gaussian kernels with different $\sigma$ values. 
    White rings in the input image denote GT out-of-plane coordinate with a ring size proportional to proximity to the displayed slice, \ie smaller rings correspond to annotations in more distant slices.
    For larger $\sigma$, out-of-plane annotations are seen to project onto the slice DM.
    (\textbf{b-e}) Comparison of different detection algorithmic choices and thresholds for the application of NMS on GT DMs (green) and DMs predicted with \emph{UNet} (red). 
    Yellow lines on the images represent the patch size of CNN output, equivalently the tiling boundary in the \emph{M}$_{\!\mathrm{conv}}$ strategy. 
    Magenta lines depict the patch size for the \emph{M}$_{\!\mathrm{peak}}$ strategy. 
    GT annotations are represented as rings ($\circ$), and predictions as crosses ($\times$). 
    Both $\circ$ and $\times$ are green when they form a TP pair, or red for FP or FN.  
    The yellow arrows mark an example where \emph{M}$_{\!\mathrm{peak}}$ helps, and the blue arrows another where \emph{K}$_{\!\mathrm{max}}$ helps. 
    The graphs show the detection metrics as the mean (solid line) and 95\% confidence interval across samples (n=7) in the test set across detection thresholds, for each tested experimental setting and $\sigma$ value. 
    For the upper bound GT cases in (b-d), one can see in the results (\eg in F1-score) that the proposed (\emph{K}$_{\!\mathrm{max}}$\&\emph{M}$_{\!\mathrm{peak}}$) method is invariant to large ranges of $\sigma$ and threshold parametrization.
    }
    \label{fig:dm}
\end{figure}

While DMs are used as a suitable framework for cell detection, their design specifics vary considerably. 
DMs encode cells as a local kernel with a peak at its centre and monotonically decreasing outwards. 
We specifically employ a 3D Gaussian kernel with a standard deviation $\sigma$ (illustrated in Fig.\ \ref{fig:dm}a), the nearly flat peak of which is advantageous for learning from GT annotations that may not be precise. 
In this section, we propose the use of two DM design strategies, which are first evaluated on GT DMs for comparisons to remain agnostic to chosen DM prediction methods.

Traditional tiling methods for image segmentation utilize input image patches larger than expected prediction output in order to take into account the pixels lost in convolutional layers~\cite{Gomariz2020, gomez2019deepimagej}, which we refer to as convolutional margin (\emph{M}$_{\!\mathrm{conv}}$). 
In the context of detection, \emph{M}$_{\!\mathrm{conv}}$ often leads to detection of multiple duplicated cells in neighbouring patches, which need to be somehow combined, often leading to a lower detection precision (Fig.\ \ref{fig:dm}c).
This effect is accentuated for higher values of $\sigma$ and lower DM thresholds, since these factors increase the chances of localizing coordinates at patch borders, where the actual peak may lie within a neighbouring patch.
The \emph{M}$_{\!\mathrm{peak}}$ tiling strategy we propose employs a smaller output patch (magenta), which is obtained by cropping the CNN output (yellow) with a small distance to ensure that duplicated detections falling within that margin are considered only in one of the patches (yellow arrow in Fig.\ \ref{fig:dm}b,c).

The second strategy (\emph{K}$_{\!\mathrm{max}}$) is designed to address a problem arising in the GT DM generation by using the process of kernel sum (\emph{K}$_{\!\mathrm{sum}}$), an approach common in earlier works~\cite{falk2019, Xie2018}. 
Adding kernels with \emph{K}$_{\!\mathrm{sum}}$ from close-by coordinates may erroneously result in single peaks.
In addition, \emph{K}$_{\!\mathrm{sum}}$ artificially increases the dynamic range of DMs complicating CNN predictions and confounding peak density with prediction strength (nearby peaks leading to relatively larger GT DM responses).
This problem is illustrated in Supplementary Fig.\ \ref{suppfig:gaussian_aggregation}, and marked with a blue arrow in an example image in Fig.\ \ref{fig:dm}d.
With \emph{K}$_{\!\mathrm{max}}$, Gaussians are compounded by their maximum, as recently proposed in a different experimental setting~\cite{Youn2020}, hence facilitating their separation by interpreting densities as an intersection rather than an accumulation, and eliminating the sensitivity of detection metrics to the spatial spread of a kernel, \ie $\sigma$ of Gaussian.

Careful selection of $\sigma$ and detection threshold values can produce satisfactory detection results on GT DMs with either of the traditional \emph{M}$_{\!\mathrm{conv}}$ or \emph{K}$_{\!\mathrm{sum}}$ strategies.
However, the insensitivity of our proposed method (\emph{K}$_{\!\mathrm{max}}$\,\&\,\emph{M}$_{\!\mathrm{peak}}$) to both parameters in the DM design make it more suitable for detection on predicted DMs, since application of trained CNNs is expected to produce DM values and shapes differently to those observed for GT DMs. 
We subsequently train a \emph{UNet} model for prediction of such DMs with the results shown in Fig.\ \ref{fig:dm}e, where detection quality is observed to highly depend on the detection threshold. 
While too small or too high thresholds are consistently worse, a high variation is observed across evaluated test samples, contrary to the clearer patterns observed for detection on GT DMs.

\subsection*{Probabilistic classification of cell proposals}
Application of NMS with a DM-based threshold hinders cell detection by ($i$)~introducing a hyperparameter that, as shown in the previous section, highly influences the detection quality and ($ii$)~preventing a probabilistic interpretation of the predictions that can be linked to their confidence.
We hereby propose an alternative that addresses these drawbacks without compromising the detection quality. 
First, we generate numerous cell proposals by using NMS on predicted DMs without applying a threshold (effectively setting the threshold to 0), which guarantees high recall but low precision because all possible peaks are considered as cells, as shown in Fig.\ \ref{fig:dm}e.
Second, we aggregate summary statistics from the neighbourhood of each proposal as a feature vector. 
These features act as input to a binary classifier that, in contrast to the deterministic threshold previously employed, assigns each of the proposals a probability of being a cell.
Our method hence provides a mapping from the regressed DM to a probability space for each identified cell.

\begin{figure}[t]
    \centering
    \includegraphics[width=0.75\textwidth]{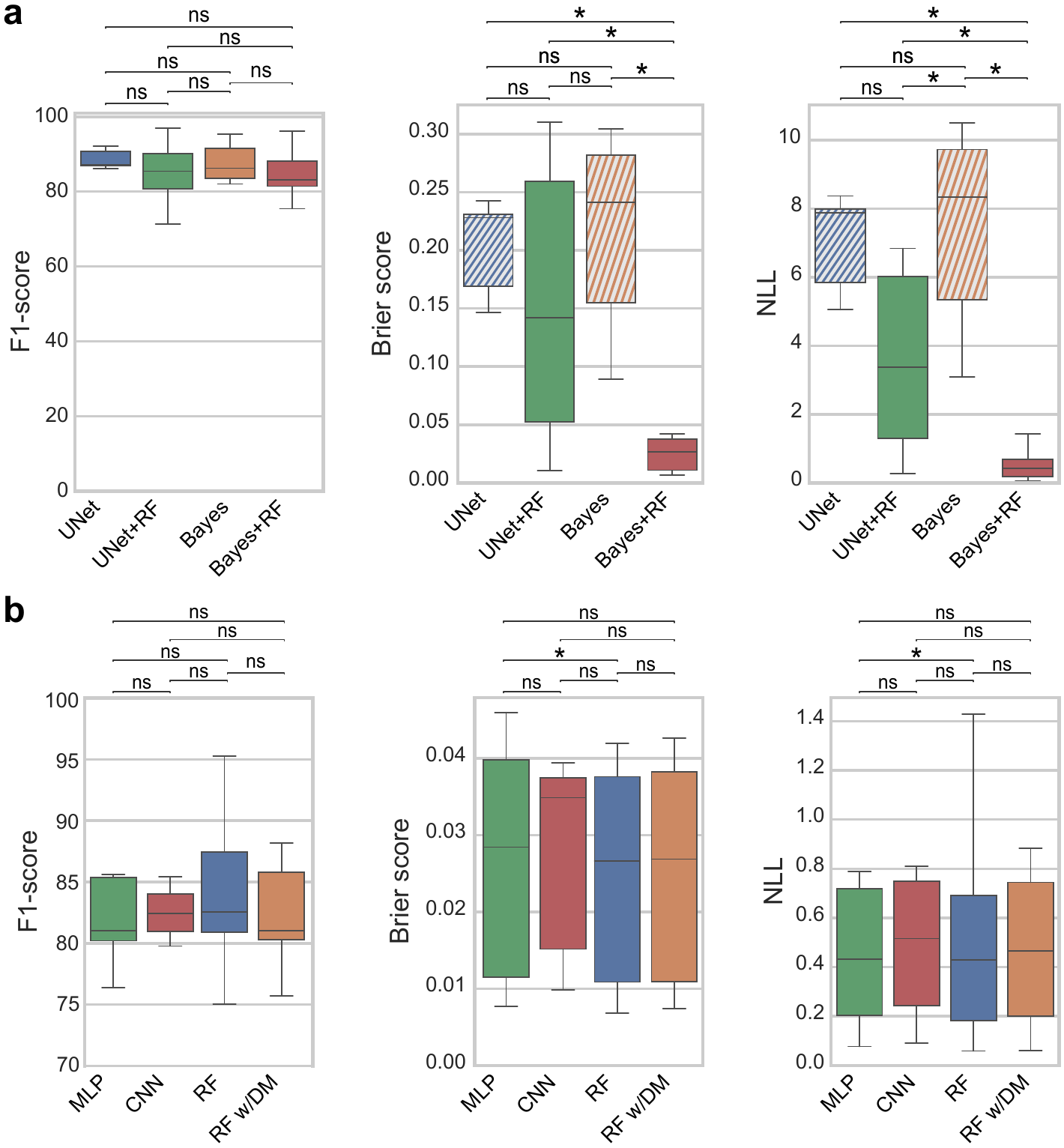}
    \caption{Prediction metrics for the different proposed cell detection models.
    (\textbf{a}) Comparison between \emph{UNet} and its Bayesian implementation \emph{Bayes} when selecting the DM threshold from the validation set, and the respective models when adding a probabilistic binary classifier RF (\emph{+RF}).
    Brier score and NLL evaluate probabilistic predictions (lower is better). 
    Since threshold-based models are deterministic, they are evaluated by assigning a fixed probability of 1 to positive predictions (TP or FP), and 0 to FN (shown in stripes to indicate this evaluation convention). 
    (\textbf{b}) Comparison of different probabilistic classifier strategies added to \emph{Bayesian UNet}. 
    All models are evaluated by 4-fold cross-validation on the test set (n=7).
    Significance is indicated with p-value$\leq$0.05~(*).
    }
    \label{fig:metrics_preds}
\end{figure}

As the binary classifier, we initially propose the use of a Random Forest (RF)~\cite{Breiman2001}. RFs have been described to be robust against overfitting, produce well-calibrated results, and provide accurate predictions when input features contain representative descriptions of the output labels~\cite{Breiman2001, gomariz2021uncertaintymarkers}. 
This latter characteristic is particularly suitable for our task, as we expect our predicted DMs to display information for easy discrimination of cell instances. 
In Fig.\ \ref{fig:metrics_preds}a we compare this RF classifier added onto the above baseline DM estimator, denoted as \emph{\mbox{UNet+RF}}, with the baseline \emph{UNet} where the DM threshold is selected from a validation set (details in Section \textit{Methods - \nameref{sec:training_evaluation}}). 
The results show that our proposed strategy can effectively remove the need for the said detection threshold, without sacrificing detection accuracy, \ie significantly affecting the $F1$-score.

Brier score~\cite{brier1950verification} and negative log-likelihood (NLL) are scoring metrics employed for the evaluation of calibration quality (lower is better) of probabilistic predictions~\cite{Gneiting2007, Guo2017}. 
These scores indicate that \emph{\mbox{UNet+RF}} leads to a slight yet not significant improvement in calibration over \emph{UNet}.

\subsection*{Bayesian regression networks for calibrated detections}
Motivated by the advantages in calibration with the use of different Bayesian DL techniques reported for different domains~\cite{kendall2017uncertainties, epistemicprobs_hiasa2019automated, Mehrtash2020, gomariz2021uncertaintymarkers}, we adapt our framework to take uncertainties in DM regression into account, which we denote as \emph{Bayes} and describe in \textit{Methods - \nameref{sec:bayesian}}.
This method produces, in addition to a predicted DM, also spatial epistemic and aleatoric uncertainty maps, which can also be provided to the classifier (RF) as additional input.
While the Bayesian detection accuracy (F1-score) does not differ significantly from the \emph{UNet} results, either with or without a probabilistic classifier (RF), it is seen in Fig.\ \ref{fig:metrics_preds}a that our proposed classifier combination with a Bayesian network (\emph{\mbox{Bayes+RF}}) substantially improves the calibration quality seen in Brier score and NLL.

Although we employ RF due to its advantages described in the previous section, we show in Fig.\ \ref{fig:metrics_preds}b that replacing it with other binary classifiers, namely a multilayer perceptron (\emph{MLP}) or a \emph{CNN}, produces similar results. 
Omitting uncertainties and employing RF only on features created from the DM (\emph{RF w/DM}) does not entail deterioration of detection quality nor calibration, implying that for Bayesian networks the prediction DMs alone hold the information necessary for inferring prediction probability.
However, as seen in Fig.\ \ref{fig:metrics_preds}a, the Bayesian output alone (\emph{Bayes}) is not well calibrated, and has an even lower Brier score and NLL than \emph{UNet+RF}.
These results indicate that the use of our binary classifier is key to obtain accurate confidence estimations.
But even if the processing of uncertainties is not fundamental, this classifier truly excels in calibration when employed on Bayesian predictions (\emph{Bayes+RF} or \emph{RF w/DM}), where the Brier score and NLL are significantly lower than for \emph{UNet+RF}.
Overall, the best results in detection $F1$-score and calibration (as measured by Brier score and NLL) are achieved with \emph{Bayes+RF} on feature vectors including uncertainty statistics, a method setting which we use for further experiments.

\subsection*{Qualitative assessment of detection proposals}
\begin{figure}[tb!]
    \centering
    \includegraphics[width=\textwidth]{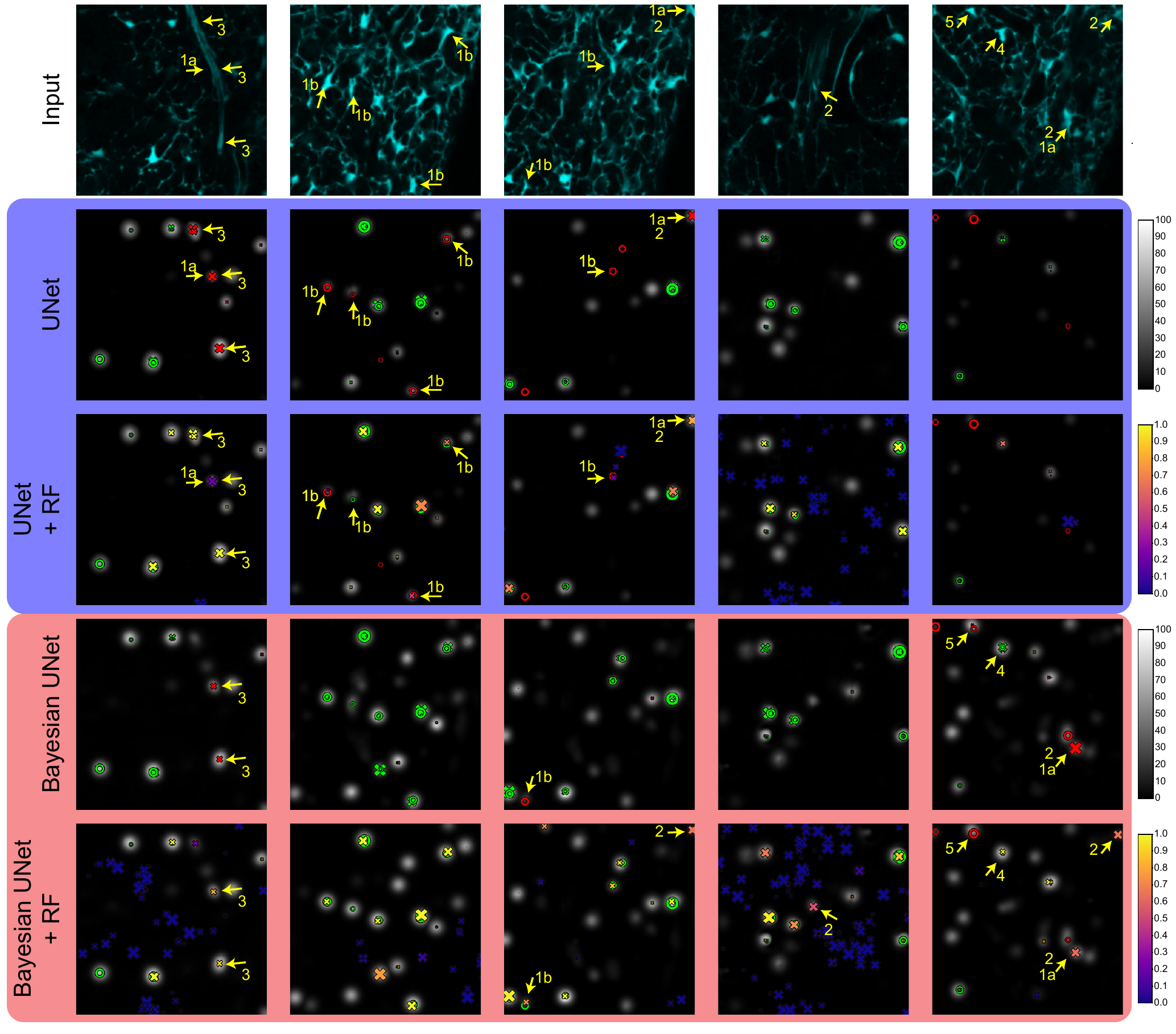}
    \caption{Visualization of predictions with different proposed models on example 2D slices of the test set.
    GT coordinates are represented by rings ($\circ$) and predictions by crosses ($\times$), with a size proportional to their proximity to the displayed slice. 
    In deterministic models (\emph{UNet} and \emph{Bayes}) coordinates are coloured green when they have a positive match (TP) and red when they do not (FP for $\times$, FN for $\circ$), according to Hungarian matching. 
    In probabilistic models (\emph{\mbox{UNet+RF}} and \emph{\mbox{Bayes+RF}}) the predictions are coloured according to their associated probability (colorbar at the right), whereas GT annotations follow the deterministic colouring scheme, considering positive predictions as those with $p\geq0.5$.
    Arrows mark specific examples discussed in the text according to their accompanying number.
    }
    \label{fig:preds_illus}
\end{figure}

To better understand the behaviour of our presented deterministic and probabilistic models we include in Fig.\ \ref{fig:preds_illus} visualizations of example predictions on the test set, with relevant findings marked with numbered arrows. 

Examples marked with arrows \textbf{1} show the benefits of probabilistic models on several predictions that are mistaken in their deterministic counterpart, either by assigning probabilities below 1 to FP~(\textbf{1a}) or above 0 for FN~(\textbf{1b}).
While in the best cases these predictions are turned respectively into TN ($p<0.5$) or TP ($p\geq0.5$), simply accounting for their probabilities lowers the confidence of wrong predictions, a benefit confirmed by the lower Brier score and NLL of probabilistic models (Fig.\ \ref{fig:metrics_preds}a). 

We also report a number of observed failure cases.
Some FP marked with arrows \textbf{2} appear as TP given the input image, and these may indeed have been mistakenly overlooked during GT annotation.
Interestingly, all such observed examples have a relatively low prediction confidence ($0.5 \leq p < 0.75$), potentially explaining or emulating the annotator's oversight. 
FP marked with arrows \textbf{3} are due to the presence of an artery: a structure unobserved in the training set that displays high intensities even in the absence of cells.
Other errors are due to the prediction of large density blobs, either producing multiple predictions for a single GT~(arrow \textbf{4}) or producing a local maximum too far from its respective GT~(arrow \textbf{5}).

\subsection*{Probabilistic spatial characterization of bone marrow stromal cells with calibrated cell detection}
\label{sec:bmquant}

\begin{figure}[bt]
    \centering
    \includegraphics[width=1\textwidth]{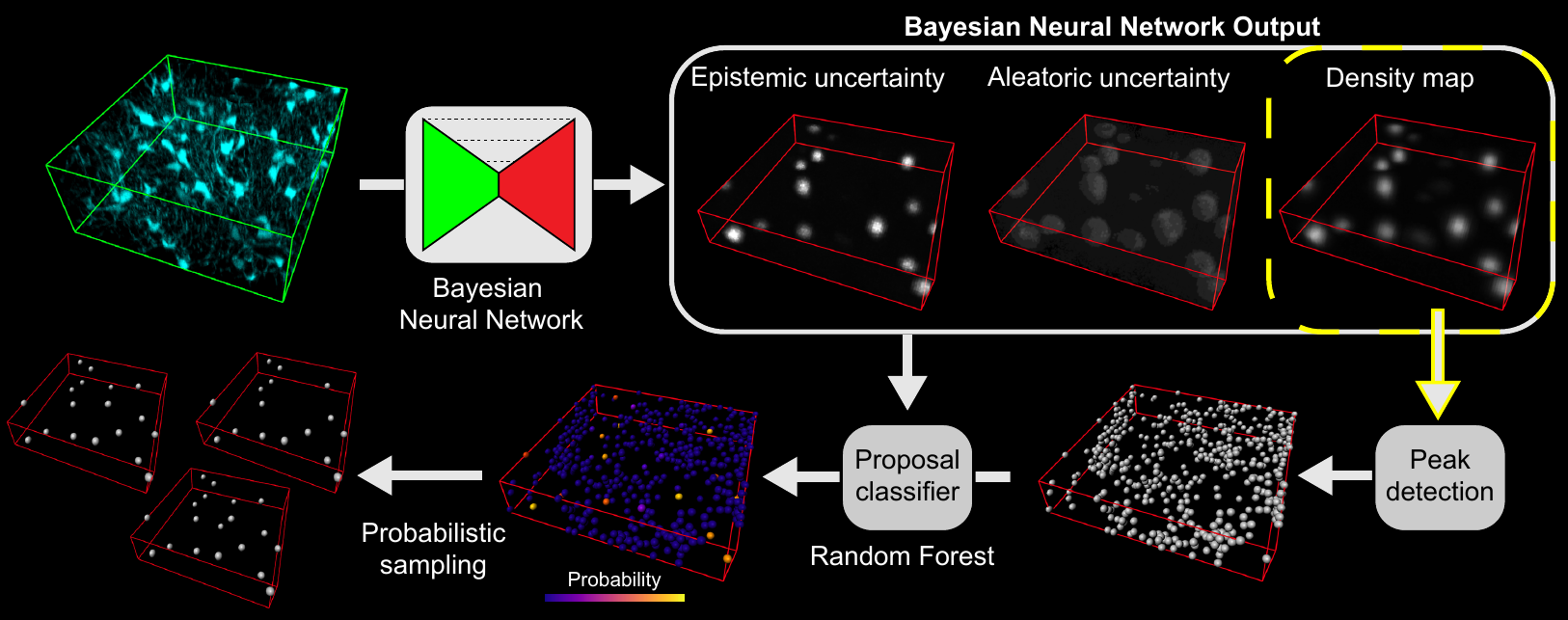}
    \caption{Representation of our proposed method for probabilistic cell detection.
    Input patches are processed by a Bayesian CNN (\emph{Bayes}) to regress an output DM and its corresponding uncertainties. 
    These outputs are employed by a classifier (\emph{RF}) to assign a probability or confidence to a large number of cells proposed by application of peak detection on the DM. 
    The resulting proposals can be sampled according to their assigned probability.
    }
    \label{fig:pipeline_sampling}
\end{figure}

\begin{figure}[bt!]
    \centering
    \includegraphics[width=\textwidth]{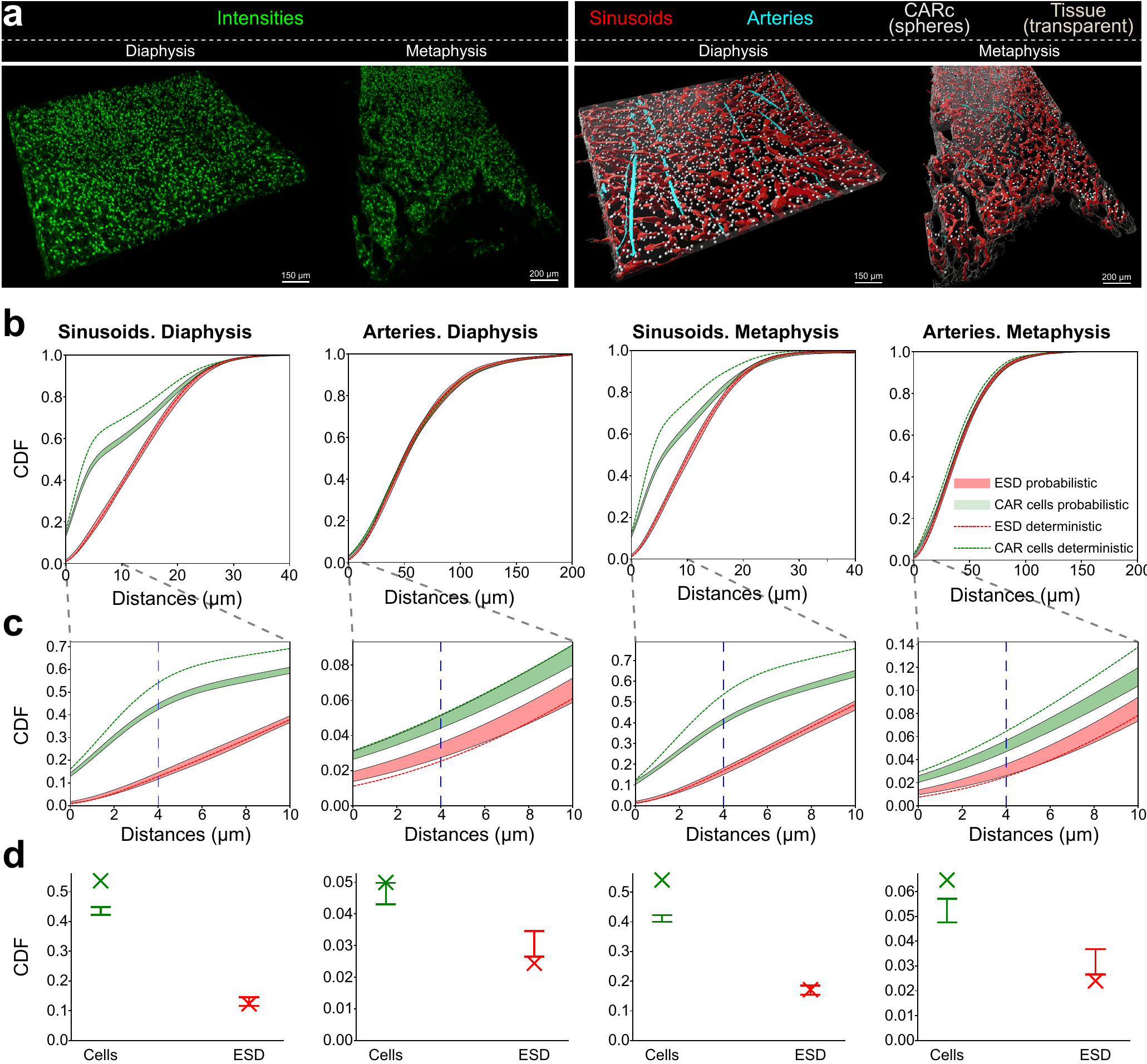}
    \caption{Quantification of spatial distributions in the BM stroma with our methods producing calibrated cell detections.
    (\textbf{a})~Example visualizations of BM diaphysis and metaphysis regions, with the intensities employed for detection of CAR cells (top) and the computational representation (segmentation and detection) of the different structures employed in the analysis (bottom). 
    (\textbf{b})~CDF of distances to sinusoids and arteries for both BM regions, zoomed-in in \textbf{c} to highlight the confidence ranges of the probabilistic analysis.
    Dashed lines are employed for \emph{deterministic} results.
    \emph{Probabilistic} results are represented as envelopes containing the maximum and minimum CDF values for each distance (n=50). 
    (\textbf{d})~Cross section of the CDF at a distance of \SI{4}{\micro\meter} (dashed blue line in \textbf{c}) to emphasize the differences between probabilistic measures capturing a confidence range (depicted as \rotatebox[origin=c]{90}{$|$\!---\!$|$}) with their deterministic counterpart that calculates a single value (depicted as $\times$). 
    }
    \label{fig:sample_analysis}
\end{figure}

In this section we deploy our calibrated cell detection framework (illustrated in Fig.\ \ref{fig:pipeline_sampling}) to revise the quantification of CAR cell spatial distributions in the context of BM stroma presented in~\cite{gomariz2018quantitative}, where an analysis of densities of this cell type and its spatial associations were evaluated with deterministic conventional image processing (CIP) methods that limited the statistical power of the observations. 
The use of spatial point processes was proposed to describe patterns with cumulative distribution functions (CDFs) of empty space distances (ESDs) and distances from CAR cells, which we apply here (details in \textit{Methods - \nameref{sec:bmquant_methods}}) to describe their distribution relative to sinusoids and arteries segmented in a separate study~\cite{Gomariz2020} for diaphysis and metaphysis, two distinct BM regions shown in~Fig.\ \ref{fig:sample_analysis}a.

We propose two different types of analysis, for which the  probabilities from \emph{\mbox{Bayes+RF}} predictions are treated differently. 
A \emph{deterministic} result is extracted for each BM sample by analyzing only cells predicted with $p \geq 0.5$, which contribute equally to the analysis regardless of their prediction confidence.
Alternative results are extracted with a \emph{probabilistic} strategy that repeats the experiment multiple times (herein 50), each time sampling cell proposals with a probability given their predicted $p$.
The contribution of each individual cell to this \emph{probabilistic} analysis hence becomes proportional to its associated confidence, which we hypothesize to statistically be more rigorous than the \emph{deterministic} interpretation; as also evaluated with the experiments below.

A comparison of the observations reported with CIP  in~\cite{gomariz2018quantitative} with the \emph{deterministic} and \emph{probabilistic} analyses hereby presented is shown in Table~\ref{tab:bmquant}.
The deterministic interpretation of the outputs achieved with \emph{Bayes+RF} already increases the statistical power of the previous quantification with CIP, as our presented method allows for automation, applicability to more samples, and better detection metrics. 
Application of our \emph{probabilistic} analysis reveals that, relative to the \emph{deterministic} results, there is an increase in the global density of cells and a decrease of them in the vicinity of both sinusoids and arteries. 
The \emph{probabilistic} interpretation of ESD implies random sampling of the empirical values, hence the similarities to \emph{deterministic} results in \emph{\%Volume} and between ESD CDFs. 
Although a sample-wise aggregation of the values is included for comparison with earlier works, we also aggregate them for the entire dataset as a final quantitative report emphasizing that only a \emph{probabilistic} analysis accounts for the uncertainty of the predictions. 

\begin{table}[!bt]
    \centering
    \caption{Analysis results on bone marrow stroma dataset as previously reported with CIP and the \emph{deterministic} and \emph{probabilistic} models proposed herein. 
    Adjacency refers to distances smaller than \SI{4}{\micro\meter}.
    Standard deviation (SD) is measured across samples in \emph{sample-wise} rows, and across prediction replicates (n=50) in \emph{aggregated dataset} rows.  
    All results are shown as mean$\pm$SD. 
    The reported detection metrics have a deterministic interpretation only, and are hence similar for both deterministic and probabilistic analyses. 
    Nevertheless, the calibration metrics indicate that our probabilistic method estimates much better the chances of a result being accurate.
    More details on these metrics can be found in \textit{Methods - \nameref{sec:training_evaluation}}. 
    }
    \resizebox{\textwidth}{!}{
    \begin{tabular}{c|l|cc|cc|cc|}
    \cline{2-8}
     & \textbf{Method} & \multicolumn{2}{c|}{ \textbf{CIP}~\cite{gomariz2018quantitative} } & \multicolumn{2}{c|}{ \textbf{Deterministic} } & \multicolumn{2}{c|}{ \textbf{Probabilistic} } \\
    \cline{2-8}
    & \textbf{Region} & Diaphysis & Metaphysis & Diaphysis & Metaphysis & Diaphysis & Metaphysis \\
    \hline
    \multicolumn{1}{|c|}{\multirow{5}{*}{\textbf{\makecell{Sample-\\wise\\analysis}}}} & 
    Density (cells/mm$^3$ x$10^4$) & 3.78\footnotesize{$\pm$0.14} & 3.21\footnotesize{$\pm$0.14} & 3.02\footnotesize{$\pm$0.32} & 3.05\footnotesize{$\pm$0.47} & 3.30\footnotesize{$\pm$0.23} & 3.41\footnotesize{$\pm$0.27} \\
    \multicolumn{1}{|c|}{} & \% Cells adjacent to sinusoids & 64.0\footnotesize{$\pm$0.7} & n/a & 50.71\footnotesize{$\pm$5.78} & 51.46\footnotesize{$\pm$3.92} & 42.89\footnotesize{$\pm$4.56} & 43.94\footnotesize{$\pm$2.60} \\
    \multicolumn{1}{|c|}{} & \% Volume adjacent to sinusoids & n/a & n/a & 11.71\footnotesize{$\pm$1.15} & 14.46\footnotesize{$\pm$2.74} & 11.73\footnotesize{$\pm$1.71} & 14.45\footnotesize{$\pm$2.74} \\
    \multicolumn{1}{|c|}{} & \% Cells adjacent to arteries & n/a & n/a & 5.10\footnotesize{$\pm$4.30} & 3.58\footnotesize{$\pm$1.48} & 4.53\footnotesize{$\pm$3.29} & 3.14\footnotesize{$\pm$1.24} \\
    \multicolumn{1}{|c|}{} & \% Volume adjacent to arteries & n/a & n/a & 1.21\footnotesize{$\pm$0.65} & 1.19\footnotesize{$\pm$0.49} & 1.24\footnotesize{$\pm$0.79} & 1.18\footnotesize{$\pm$0.50} \\
    \hline
    \multicolumn{1}{|c|}{\multirow{5}{*}{\textbf{\makecell{Aggregated\\dataset\\analysis}}}}
    & Density (cells/mm$^3$ x$10^4$) & n/a & n/a & 3.07 & 2.99 & 3.28\footnotesize{$\pm$0.01} & 3.41\footnotesize{$\pm$0.01} \\
    \multicolumn{1}{|c|}{} & \% Cells adjacent to sinusoids & n/a & n/a & 50.23 & 50.78 & 42.97\footnotesize{$\pm$0.14} & 43.06\footnotesize{$\pm$0.17} \\
    \multicolumn{1}{|c|}{} & \% Volume adjacent to sinusoids & n/a & n/a & 14.93 & 11.37 & 11.38\footnotesize{$\pm$0.14} & 14.92\footnotesize{$\pm$0.23} \\
    \multicolumn{1}{|c|}{} & \% Cells adjacent to arteries & n/a & n/a & 4.78 & 4.26 & 4.32\footnotesize{$\pm$0.07} & 3.74\footnotesize{$\pm$0.11} \\
    \multicolumn{1}{|c|}{} & \% Volume adjacent to arteries & n/a & n/a & 1.31 & 1.39 & 1.31\footnotesize{$\pm$0.06} & 1.36\footnotesize{$\pm$0.11} \\
    \hline
    \multicolumn{1}{|c|}{\multirow{2}{*}{\textbf{\makecell{Throughput}}}} & \# samples (used for arteries) & 6 (0) & 6 (0) & 33 (24) & 10 (6) & 33 (24) & 10 (6) \\
    \multicolumn{1}{|c|}{} & Manual work & \multicolumn{2}{c|}{ 15 min / sample } & \multicolumn{2}{c|}{ None (fully automatic) } & \multicolumn{2}{c|}{ None (fully automatic) } \\
    \hline
    \multicolumn{1}{|c|}{\multirow{3}{*}{\textbf{\makecell{Detection \\metrics}} }} & $F1$-score & \multicolumn{2}{c|}{ 74.76\footnotesize{$\pm$9.63} } & \multicolumn{4}{c|}{ 84.84\footnotesize{$\pm$6.73}} \\
    \multicolumn{1}{|c|}{} & Precision & \multicolumn{2}{c|}{ 73.49\footnotesize{$\pm$21.77} } & \multicolumn{4}{c|}{ 83.84\footnotesize{$\pm$9.72} } \\
    \multicolumn{1}{|c|}{} & Recall & \multicolumn{2}{c|}{ 83.63\footnotesize{$\pm$15.38} } & \multicolumn{4}{c|}{ 86.62\footnotesize{$\pm$7.73} } \\
    \hline
    \multicolumn{1}{|c|}{\multirow{2}{*}{\textbf{\makecell{Calibration \\metrics}}}} & Brier [\%] & \multicolumn{2}{c|}{ 39.48\footnotesize{$\pm$12.56} } & \multicolumn{2}{c|}{ 25.81\footnotesize{$\pm$10.47} } & \multicolumn{2}{c|}{ 3.46\footnotesize{$\pm$3.65} } \\
    \multicolumn{1}{|c|}{} & NLL & \multicolumn{2}{c|}{ 13.64\footnotesize{$\pm$4.34} } & \multicolumn{2}{c|}{ 8.91\footnotesize{$\pm$3.62} } & \multicolumn{2}{c|}{ 0.52\footnotesize{$\pm$0.48} } \\
    \hline
\end{tabular}}
    \label{tab:bmquant}
\end{table}

The application of \emph{deterministic} analysis in the characterization of distances through their CDFs, shown in Fig.\ \ref{fig:sample_analysis}b-d, confirms the previously reported trend of the preferential localization of CAR cells in close proximity to sinusoids, both in diaphysis and metaphysis. 
Furthermore, the access to segmented arteries unavailable in~\cite{gomariz2018quantitative} allows to describe a much subtler tendency of CAR cells to also be preferentially residing within the vicinity of these specialized vascular structures.
The application of the \emph{probabilistic} analysis permits the extraction of these curves as confidence intervals (envelopes) reflecting the confidence in results when taking into account prediction probabilities. 
Strikingly, the deterministic curves lie outside these envelopes for large spatial distance ranges.
This suggests that a \emph{probabilistic} analysis is not only beneficial in accounting for uncertainty, but also in potentially revealing patterns only visible by considering confidence associated to the model's predictions. 

\section*{Discussion}
In this study we propose a DL-based cell detection framework designed for its application on 3D FM datasets that addresses several problems in previous DM regression methods for detection.
Additionally, we introduce a method for well-calibrated probabilistic predictions that can be incorporated in powerful statistical analysis frameworks. 

We first argue the importance of adapting existing tiling strategies to take a supplementary margin into account (\emph{M}$_{\!\mathrm{peak}}$), without which any detection results are inherently flawed, even upon perfect DM predictions, due to their merging at patch boundaries. 
Compounding annotations as \emph{K}$_{\!\mathrm{max}}$, a technique that, to the best of our knowledge,  had not been adopted in cell detection to date, is also found to be key for achieving accurate results without the influence of $\sigma$ in the DM design.
When predicting DMs with CNNs, the density threshold employed for proposals becomes a major limitation, as it may lead to unexpectedly inferior detection results if not diligently analyzed. 
Removing this threshold hyperparameter without compromising the detection quality is by itself a fundamental advantage of our probabilistic detection approach. 
Indeed, accounting for model uncertainties with deep Bayesian methods results in the additional benefit of better calibrated predictions, as evaluated by their superior Brier score and NLL. 
In such setting, we show that the classification of cell proposals is similarly effective with a number of different machine learning classifiers, suggesting that predicted DMs already pose a simple classification problem and that the key contribution resides in our proposed pipeline design rather than the specific classifier employed. 
A relevant question for future work is whether the number of features in the selected RF method can speed up the detection process without compromising the results. 

Although accounting for uncertainties in the regression CNN improved the calibration of the predictions, including them in the features for the probabilistic classifier did not significantly impact any evaluated metric.
This observation may imply that DM values can somehow reflect uncertainty information that is subsequently exploited by the classifier, consistent with the findings in~\cite{kendall2017uncertainties} that simply accounting for uncertainty during training of CNNs improves calibration of the results, without the need for post-processing them.
However, uncertainties did not seem to capture information about specific unseen structures (Fig.\ \ref{fig:preds_illus}) as has been the case in previous reports~\cite{kendall2017uncertainties, gomariz2021uncertaintymarkers}.
Whether this is a limitation caused by our dataset, the chosen Bayesian techniques, or the use of DM as a proxy for detection is worth investigating in future research. 
In particular, although not applied herein due to their impracticable computational complexity, deep ensembles that have produced more accurate uncertainties in other tasks~\cite{fort2020deep} may shed light on and provide improvements to our Bayesian approximations. 

We believe that our probabilistic cell detection framework offers a paradigm-shift in analysis tasks. 
Although the reported throughput and detection metrics already favor the use of the \emph{deterministic} interpretation of the outputs achieved with \emph{Bayes+RF} as compared to the previously employed CIP method~\cite{gomariz2018quantitative}, it is the superior calibration metrics that justify the proposed \emph{probabilistic} analysis as a superior option to the more common \emph{deterministic} alternative. 
As we showcase with the numbers reported for BM stroma, a \emph{deterministic} interpretation of the results does not contemplate mistakes in predictions. 
Instead, accounting for output confidences in a \emph{probabilistic} analysis entails a weighting of the contribution from each prediction to the final results by its associated probability, which, in correctly calibrated models, corresponds to their likelihood of being accurate. 
The application of this analysis for the characterization of the spatial distribution of CAR cells within the BM has confirmed our previous results and corroborate the preferential localization of this cell type in close association to sinusoidal vessels reported in~\cite{gomariz2018quantitative}. 
However, the revised results presented herein not only provide a higher statistical power and additional confidence estimates, but also reveal substantial differences in specific parameters defining spatial distributions of CAR cells.
For instance, the probabilistic interpretation indicates that the amount of CAR cells adjacent to sinusoids is 15.42$\pm$13.18\% and 14.61$\pm$9.86\% lower than in the deterministic results for the diaphysis and metaphysis respectively. 
In this way, in addition to the robust hypothesis testing techniques already employed in~\cite{gomariz2018quantitative}, we hereby take into account variability in our cell detection predictions. 
In fact, this latter variability is estimated from the uncertainty implicit to the method (epistemic) and the uncertainty arising from the observations (aleatoric), such that we account for all possible sources of error in our analysis.
Therefore, we trust that the proposed cell detection and probabilistic analysis strategies will become the basis for future studies of spatial distributions.

The effect of a probabilistic analysis in a segmentation setting (\eg sinusoids or arteries) will be a relevant question worth investigating in future work.
However, the sampling strategy proposed herein for detection coordinates would not be suitable for representing uncertainty for segmentation tasks, \ie sampling pixels cannot simulate ambiguities, \eg in extents along different dimensions, morphology, etc. 
Consequently, novel methods will need to be devised to establish connections between uncertainty maps and a consistent \emph{probabilistic} analysis. 

Although instance segmentation methods can be employed for similar analyses~\cite{Schmidt2018, weigert2020star, Stringer2020}, detection frameworks have the benefit that manual labeling of coordinates is substantially faster than individual pixels, especially in 3D data. 
Furthermore, while other detection methods employed for the prediction of bounding boxes consist of multi-stage approaches that require several post-processing steps~\cite{Zhao2019a}, we employ a DM regression approach that, in its standard form, only requires a threshold-based NMS~\cite{Xie2018}.
With our contributions, such a threshold hyperparameter is further eliminated with two separate supervised models that constitute an end-to-end framework without compromising the detection accuracy. 
The additional benefit of our method predicting well-calibrated probabilistic cell proposals is shown to offer a more comprehensive analysis enabling confidence intervals for any hypothesis testing, which we believe will be the base of future image-based quantification in different FM and other imaging datasets.

\section*{Methods}

\subsection*{Dataset and tiling strategy}
\label{sec:dataset}
We evaluate our detection strategies on BM samples from the dataset presented in~\cite{gomariz2018quantitative}. 
The CXCL12-GFP channel that expresses high intensities for CAR cells is employed as input image.
Cells are manually annotated by their central coordinates in 7 different samples with the properties in Supplementary Table~\ref{supptab:dataset}.

Since large FM volumes cannot fit in the memory of GPUs required for fast application of CNNs, we decompose each sample into a number of 3D patches following a previously reported tiling strategy for 2D images~\cite{Gomariz2020}.
This method ensures that output patches, which are smaller than input patches upon application of CNNs, can be seamlessly reconstructed.
In summary, Gaussian normalization (subtracting the mean and dividing by SD) is first applied to all samples, before resizing them to the isotropic resolution of \SI{1}{\micro\meter}/voxel and zero-padding with a margin $l_\textrm{pad}$.
Input patches $x \in \mathbb{R}^{l_\textrm{in}}$ are then extracted with an overlap of $l_\textrm{overlap}$ between them. 
Output patches $y \in \mathbb{R}^{l_\textrm{out}}$ are generated as GT DMs from the coordinates annotated within $x$ as detailed in next subsection for supervised training of CNNs. 
$l_\textrm{out}$ is determined by the pixels subtracted from $l_\textrm{in}$ by convolution operations within the regression CNN.

A final output $y_\textrm{tile} \in \mathbb{R}^{l_\textrm{out\_tile}}$ is obtained from $y$ depending on the tiling strategy employed. 
The \emph{M}$_{\!\mathrm{conv}}$ strategy uses $y_\textrm{tile} = y$, and \emph{M}$_{\!\mathrm{peak}}$ subtracts a supplementary margin of \SI{4}{\micro\meter} to avoid duplicated predictions in neighbour patches. 
$l_\textrm{overlap}$ is calculated such that the resulting $y_\textrm{tile}$ are adjacent to each other, except at the sample border, where an additional overlap is permitted to ensure completion of the entire sample volume.
The sizes employed for each of these strategies are reported in Supplementary Table~\ref{supptab:tiling}.
$l_\textrm{in}$ is selected to create input patches that are as big as possible while fitting in the GPU memory and being divisible at the different downsampling layers by accounting for the pixels lost due to the corresponding convolutional layers. 
Smaller $l_\textrm{in}$ results in a higher ratio of borders in the dataset, which can negatively affect the detection results when employing \emph{M}$_{\!\mathrm{conv}}$ tiling.
However, as shown in Fig.\ \ref{fig:dm}b, the proposed \emph{M}$_{\!\mathrm{peak}}$ strategy employed in the rest of the experiments is invariant to such border artifacts, and hence $l_\textrm{in}$ may only impact the computational speed.

For the training and evaluation of models, 20\% of the samples within the complete dataset $D$ are split as a separate test set $D_\textrm{test}$ where the different methods are evaluated. 
Of the remaining samples ($D \backslash D_\textrm{test}$), 80\% are randomly selected for inclusion in a training set $D_\textrm{train}$ employed for the training of DM regression models, and 20\% in a validation set $D_\textrm{val}$ where different hyperparameters of these models are evaluated. 
This separation into training and validation from $D \backslash D_\textrm{test}$ is repeated with the same ratio to create a parallel split $\tilde{D}_\textrm{train}$ and $\tilde{D}_\textrm{val}$, which are employed distinctly in the training of binary classifiers, as explained in the next sections.

\subsection*{Density map generation}
\label{sec:dmdesign_methods}
Let $C \in \mathbb{R}^{N_c \times 3}$ be the annotated GT locations within an image $x$, where $N_c$ is the number of annotations.
Let each location be denoted as $c \in \mathbb{R}^3$, and $G_\sigma$ the Gaussian function be defined as:
\begin{equation*}
    G_\sigma(s) = \frac{1}{\sigma \sqrt{2\pi}} \exp \left(-\frac{s^2}{2\sigma^2}\right).
\end{equation*}

Compounding $G_\sigma$ using the \emph{K}$_{\!\mathrm{sum}}$ strategy commonly employed in previous works results in a DM $y$ with the value at each pixel $k \in \mathbb{R}^3$ calculated as:
\begin{equation*}
    y_k = \sum_{c} G_\sigma(k - c) , \quad \textrm{where} \quad \lVert k - c \rVert_2  \leq l_g,
\end{equation*}
considering points only within a radius $l_g$=\SI{16}{\micro\meter} for computational efficiency, since points further away will have an insignificant contribution.

In the \emph{K}$_{\!\mathrm{max}}$ strategy, the different $G_\sigma$ associated to each of the coordinates is compounded by their maximum:
\begin{equation*}
    y_k = \max_{c} \left(\left\{G_\sigma(k - c) \right\}\right) , \quad \textrm{where} \quad \lVert k - c \rVert_2  \leq l_g.
\end{equation*}

Coordinates are localized in the resulting DMs by application of NMS with a minimum distance between peaks of \SI{4}{\micro\meter}.
This value is chosen as the average cell radius, which was also observed in the training data to be smaller than the minimum distance between any GT annotation points.

\subsection*{Neural Networks for density map regression}
\label{sec:bayesian}
The \emph{UNet} model $f_\textrm{UNet}$ employed herein is based on the one proposed in~\cite{falk2019}, which we adapt to 3D  by replacing 2D with 3D convolutional layers, with less feature channels in order to compensate for the capacity increase related to the use of 3D kernels. 
In addition, we include residual connections~\cite{He2016} for each convolutional block.
We define a convolutional block with $q$ nodes as: 
\begin{equation*}
C_q(a) = \textrm{ReLu}\bigg( \textrm{ReLu}\bigg( h_q \Big( \textrm{ReLu}\big(h_q(a)\big)\Big)\bigg) + a \bigg),    
\end{equation*}
where  ReLu is the Rectified Linear Unit~\cite{relu10}, $a$ is an activation at any level of the CNN, and $h_q$ is a 3D convolutional layer with a $3\times3\times3$ kernel and $q$ channels.

Denoting 3D max pooling layers with $2\times2\times2$ kernel as MP and upsampling layers with the same kernel as UP, we build \emph{UNet} composed of the following layers:
$C_{16} \shortrightarrow \! \text{MP} \!
\shortrightarrow\!C_{32} \shortrightarrow \! \text{MP} \!
\shortrightarrow\!C_{64} \shortrightarrow \! \text{UP} \!
\shortrightarrow\!C_{32} \shortrightarrow \! \text{UP} \!
\shortrightarrow\!C_{16} \shortrightarrow \! h_1$.
Skip connections are added to connect blocks with the same resolution as described in~\cite{falk2019}. 
This \emph{UNet} is trained with L2 loss on predictions $\hat{y}=f_\textrm{UNet}(x)$ as:
\begin{equation*}
    L = \sum_{(x_k,y_k) \in (x,y)} \left( y_k - \hat{y}_k \right) ^2.
\end{equation*}

The Bayesian UNet $f_\textrm{Bayes}$ employs the same underlying architecture as $f_\textrm{UNet}$, but different training and inference strategies, following the design choices justified in~\cite{gomariz2021uncertaintymarkers}.
The last layer is changed from $h_1$ to $h_2$ to take aleatoric uncertainty $u_a$ into account, which is calculated simultaneously with the DM as $[\hat{y}, u_a] = f_\textrm{Bayes}(x)$. 
A ReLu activation is applied to $u_a$ to ensure positive values.
The training loss function then becomes:
\begin{equation*}
    L_\textrm{Bayes} = \sum_{(x_k,y_k) \in (x,y)} \frac{\left( y_k - \hat{y}_k \right) ^2}{2 u_{a,k}} + \frac{1}{2}\log{u_{a,k}}.
\end{equation*}
In addition, MC dropout is used to take into account epistemic uncertainty $u_e$ by randomly (with a uniform probability of 0.2) deleting some of the convolutional layers $h$ except the last one~\cite{gal2015dropout}.
With the output being stochastic, 50 samples $[\hat{y}^t, u_a^t]$ are drawn at inference to calculate the final $\hat{y}$.
For this random sampling, $u_a$ is used as the respective detection mean, and $u_e$ as the pixel-wise SD across the 50 samples $\hat{y}^t$.

All CNNs in this work are implemented with TensorFlow v2.3~\cite{tensorflow2015whitepaper} and deployed on an NVIDIA GeForce GTX TITAN X GPU with 12 GB of VRAM. 
A batch size of 4 is used, which is the maximum size that fits the GPU memory available. 
All models were trained with Adam optimizer~\cite{kingma2014adam} with a learning rate of $10^{-3}$.

\subsection*{Probabilistic classification of cell proposals}
For the application of probabilistic classifiers, $\hat{N_c}$ cell proposals  $\hat{C} \in \mathbb{R}^{\hat{N_c} \times 3}$ are generated from DMs by application of a threshold-free (setting it to 0) NMS. 
A feature vector $v_c$ is then generated by extracting summary statistics for volumes with different sizes (4, 8, 16 and \SI{32}{\micro\meter}$^3$) centered on each of the $\hat{c} \in \hat{C}$ for the available predictions (DM, $u_a$, and $u_e$).
The summary statistics were heuristically selected as follows:
\begin{itemize}
    \item 5 percentiles for each of the volumes considered, which are taken uniformly in the range from the $1^\text{st}$ to the $99^\text{th}$.
    \item The ratio of voxels above 5 different thresholds $t$, selected uniformly in ranges that differ for each of the volumes considered: $t \in [1, 1.5]$ for DM, $t \in [1, 10]$ for $u_a$, and $t \in [1, 0.2]$ for $u_e$.
    \item The first 4 statistical moments: mean, SD, skewness, and kurtosis. 
\end{itemize}

Binary classification models $g$ are trained on the features $v_c$ to learn whether each proposal corresponds to a cell. 
Following the dataset notation in Section \textit{Methods} - \nameref{sec:dataset}, RF classifiers ($g_\textrm{RF}$) are trained with 128 trees and gini criterion on $\tilde{D}_\textrm{train} \cup \tilde{D}_\textrm{val}$.
The validation set is not employed separately because no hyperparameters need to be tuned. 
MLP classifiers ($g_\textrm{MLP}$) are designed with 4 hidden layers (with 50, 50, 20 and 20 nodes respectively) with ReLu activations. 
They are trained for 200 epochs with NLL loss and Adam optimizer on $\tilde{D}_\textrm{train}$.
The epoch with best accuracy on the validation set $\tilde{D}_\textrm{val}$ is selected. 

To provide and compare with a classifier alternative that is independent of the hand-crafted features  $v_c$ used above, we also used a CNN classifier $g_\textrm{CNN}$ that acts directly on a $40\times40\times40$ voxels region surrounding each proposal $\hat{c}$.
These regions are concatenated as channels from the three output volumes: DM, $u_a$ and $u_e$.
Following the notation in \textit{Methods - \nameref{sec:bayesian}}, $g_\textrm{CNN}$ is implemented with a number of convolutional blocks 
$C_q (a) = \textrm{ReLu}\left(h_q \left(\textrm{ReLu}\left(h_q \left(a\right)\right)\right)\right)$ 
and fully connected layers $\textrm{FC}_q$ with $q$ channels, each as
$C_{32} \shortrightarrow \! C_{64} \shortrightarrow \! C_{128} \shortrightarrow \! \textrm{FC}_{2048} \shortrightarrow \! \textrm{FC}_{2048} \shortrightarrow \! \textrm{FC}_{1}$.
Similarly to MLP above, this CNN classifier is also trained on $\tilde{D}_\textrm{train}$. 
We used Adam optimizer with a learning rate of $10^{-4}$, and NLL loss for 20 epochs.
The epoch with the best accuracy on $\tilde{D}_\textrm{val}$ is then selected. 

These binary models output values $p$ in the range $[0, 1]$ that can be interpreted as probabilities. 
Therefore, we consider as positive predictions those with $p \geq 0.5$.
In the case of baseline threshold-based detections (\emph{UNet} and \emph{Bayes} in Fig.\ \ref{fig:metrics_preds}), the threshold is selected on the validation set $D_\textrm{val}$ employed in the training of the regression CNNs.
Hence, we also study the effect of selecting a threshold probability $p$ from the validation set $\tilde{D}_\textrm{val}$ for the RF binary classifier employed.
Supplementary Fig.\ \ref{suppfig:validation_prob} shows that in this case, RF (denoted as $\textrm{RF}(p\leftarrow \tilde{D}_{\textrm{val}})$) performs similarly in detection $F1$-score to the RF employed in the rest of this work, which, as explained above, does not employ a separate validation set.

$g_\textrm{RF}$ and $g_\textrm{MLP}$ classifiers are implemented with scikit-learn v0.23~\cite{Pedregosa2011}, whereas $g_\textrm{CNN}$ is implemented with TensorFlow v2.3~\cite{tensorflow2015whitepaper}.

\subsection*{Evaluation details}
\label{sec:training_evaluation}

Coordinates from patches are reconstructed in the context of samples by reverting the tiling strategy in \textit{Methods - \nameref{sec:dataset}} prior to their evaluation. 
Subsequently, a pairing function $\lambda$ is defined that matches each GT coordinate $c$ to a predicted coordinate $\hat{c} = \lambda(c)$ by optimizing the linear sum assignment of their respective Euclidean distances:
\begin{equation*}
    \min_{\lambda} \sum_{c \in C} \lVert c - \lambda(c) \rVert_2\,.
\end{equation*}
This equation is solved with the Hungarian algorithm~\cite{Kuhn1955}, with some example cases illustrated in Supplementary Fig.\,\ref{suppfig:hungarian_matching}.

A TP is considered for pairs $\lVert c - \lambda(c) \rVert_2 \leq t_\textrm{match}$, where $t_\textrm{match}$ is a distance threshold.
We set $t_\textrm{match}$ to \SI{4}{\micro\meter}, the average expected cell radius and the same parameter employed as the minimum distance separating peaks in NMS (see above in Section \textit{Methods - \nameref{sec:dmdesign_methods}}). 
Assignments $\lVert c - \lambda(c) \rVert_2 > t_\textrm{match}$ produce a FP for the prediction and a FN for the GT.
A FN is counted when no prediction is assigned to a GT ($\lambda(c) = \emptyset$), and a FP when no GT is assigned to a prediction ($\lambda^{-1}(\hat{c}) = \emptyset$).
From this strategy, we calculate the Precision, Recall, and $F1$-score as:
\begin{equation*}
    \mathrm{Precision} = \frac{TP}{TP + FP}\,,
\end{equation*}
\begin{equation*}
    \mathrm{Recall} = \frac{TP}{TP + FN}\,,
\end{equation*}
\begin{equation*}
    \mathrm{\operatorname{F1-score}} = 2\cdot\frac{\mathrm{Precision} \cdot \mathrm{Recall}}{\mathrm{Precision} + \mathrm{Recall}}\,.
\end{equation*}

The calibration of the classification models is evaluated by taking into account the probabilities of predicted coordinates $p(\hat{c})$ estimated by the classifiers described in the previous section.
Coordinates predicted by deterministic models are assigned $p(\hat{c})=1$ for TP and FP, and $p(\hat{c}) = p(\lambda(c))=0$ for FN. 
Although GT coordinates do not strictly have an associated probability, we denote them as such for convenience. 
Therefore, annotated GT coordinates have $p(c)=1$, and FP are counted as $p(c) = p(\lambda^{-1}(\hat{c}))=0$. 
Calibration is assessed with the Brier score and NLL as follows:
\begin{equation*}
    \text{Brier score} = \frac{1}{N_c} \sum_{c \in C} \Big(p(c) - p\big(\lambda(c)\big)\Big)^2\,,
\end{equation*}
\begin{equation*}
    \text{NLL} = -\frac{1}{N_c}\sum_{c \in C} \bigg(p(c) \cdot \log\Big(p\big(\lambda(c)\big)\Big) + \big(1 - p(c)\big) \cdot \log\Big(1 - p\big(\lambda(c)\big) \Big)\bigg)\,.
\end{equation*}

All models were trained for 200 epochs on the training set $D_\textrm{train}$ and the epoch with the highest $F1$-score on the validation set $D_\textrm{val}$ is selected for posterior evaluation on the test set $D_\textrm{test}$. 
Results are calculated for each sample in the test set, following a 4-fold cross-validation strategy.

\subsection*{Pipeline for quantification of bone marrow stroma}
\label{sec:bmquant_methods}
The quantification of CAR cells described in Section \textit{Results - \nameref{sec:bmquant}} is performed on an extension of the dataset provided in~\cite{gomariz2018quantitative}, which includes samples that the CIP methods employed in the original work could not successfully analyze. 
Note that manually labeled samples employed for the evaluation of the detection methods form a subset of this data. 
This dataset was also employed in~\cite{Gomariz2020} using CNNs to segment sinusoids, arteries, and a tissue mask, which we utilize in our current analysis.
Sinusoids and arteries are used herein as reference structures to evaluate potential spatial distributions of CAR cells relative to them. 
The tissue mask defines the voxels inside the specimen where analysis must occur, such that out-of-tissue empty regions are ignored in spatial analysis. 

As proposed in~\cite{gomariz2018quantitative}, cellular distributions are analyzed by employing CDFs of distances to segmented structures.
The ESD aggregates distances from all background voxels to their respective closest foreground voxel (herein sinusoids or arteries) and its CDF describes the proportion of volume left for cells to distribute at different distances. 
We compare distances from predicted cell locations to a segmented structure using the ESD of the latter in CDF form.
This comparison allows for statistical testing of patterns in the localization of cells relative to the ESD.
Namely, an attraction pattern is suggested when the CDF of distances from cells is above that of a baseline ESD; and conversely an avoidance pattern, when it is below. 

Analysis is performed on a subset of cells $\tilde{C}$ obtained from the predictions $\hat{C}$ proposed by a trained \emph{Bayes+RF}. 
In a \emph{deterministic} analysis, all predictions with probabilities of at least 0.5 are considered:
\begin{equation*}
    \tilde{C} = \{\hat{c} \; | \; p(\hat{c}) \geq 0.5,  \; \forall \hat{c} \in \hat{C} \}.
\end{equation*}
In \emph{probabilistic} analyses, coordinates $\hat{C}$ are sampled according to their probabilities:
\begin{equation*}
    \tilde{C} = \{\tilde{c} \sim p(\hat{c}), \; \forall \hat{c} \in \hat{C} \}.
\end{equation*}
To incorporate the effect of cell predictions with different probabilities, the above analysis is repeated multiple times (herein $T=50$), which according to a Monte-Carlo interpretation of simulations~\cite{Baddeley_2015} permits for testing differences with a significance level $\alpha=2/(T+1)=0.04$. 
Each \emph{probabilistic} ESD is calculated by random sampling from the original ESD of a number voxels $w$ calculated from a Poisson distribution as $w \sim \operatorname{Pois}(\tilde{N_c})$, as described in~\cite{Baddeley_2015}.
CDF envelopes in this setting are created as in~\cite{gomariz2018quantitative} by ($i$) estimating the probability density function of the distances with Gaussian kernels and Scott method~\cite{Scott1992}, ($ii$) evaluating the CDF at the same distance values for all replicates, and ($iii$) employing the maximum and minimum values at each evaluated distance as the upper and lower envelope lines respectively. 

\subsection*{Statistical tests}
The two-sided Wilcoxon signed-rank test, which is non-parametric to avoid any assumption of normal distribution,  is employed to assess differences between cell detection results in a paired manner between samples.   
The 2-sample Kolmogorov-Smirnov test is employed to compare differences in distance distributions when assessing spatial patterns.

\newpage
\bibliographystyle{naturemag}  
\bibliography{bibliography.bib}

\newpage
\section*{Supplementary Materials}

\begin{suppfigure}[hbt!]
    \centering
    \includegraphics[width=\textwidth]{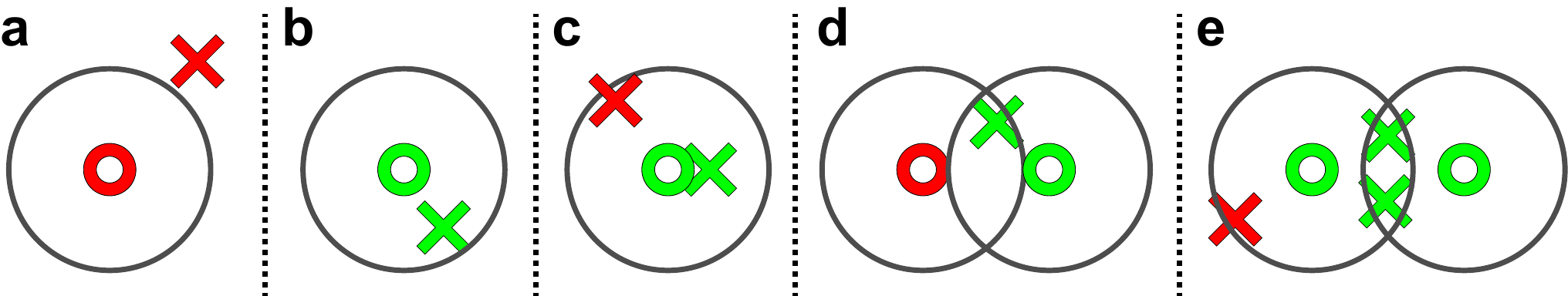}
    \caption{ 
    Illustration of Hungarian matching between predicted ($\times$) and GT ($\circ$) coordinates in different scenarios. 
    Every GT annotation ($\circ$) is paired as a positive match (TP paired, colored in green) with one and only one prediction point ($\times$) that lies within a predefined (\eg cell) radius, denoted by the gray circles. 
    Negative matches are coloured in red, and denote FN for $\circ$, and FP for $\times$. 
    (\textbf{a})~If predictions are outside the threshold distance of their corresponding GT, they are counted as FN and FP respectively.
    (\textbf{b})~If the distance is below the threshold, both are counted as a single TP.
    (\textbf{c-e})~Multiple potential matches are resolved by minimizing the linear sum assignment, \ie Hungarian matching, \eg 
    (\textbf{c})~Only the closest prediction counts as TP, while the others as FP, if multiple are within the distance threshold.
    (\textbf{d})~If the prediction is within the distance threshold to two different GT, only the closest one counts as TP, while the other becomes a FN.
    (\textbf{e})~With multiple predictions within the range of two different GT points, the further one counts as FP, while the other two form TP pairs. 
    }
    \label{suppfig:hungarian_matching}
\end{suppfigure}

\begin{suppfigure}[hbt!]
    \centering
    \includegraphics[width=\textwidth]{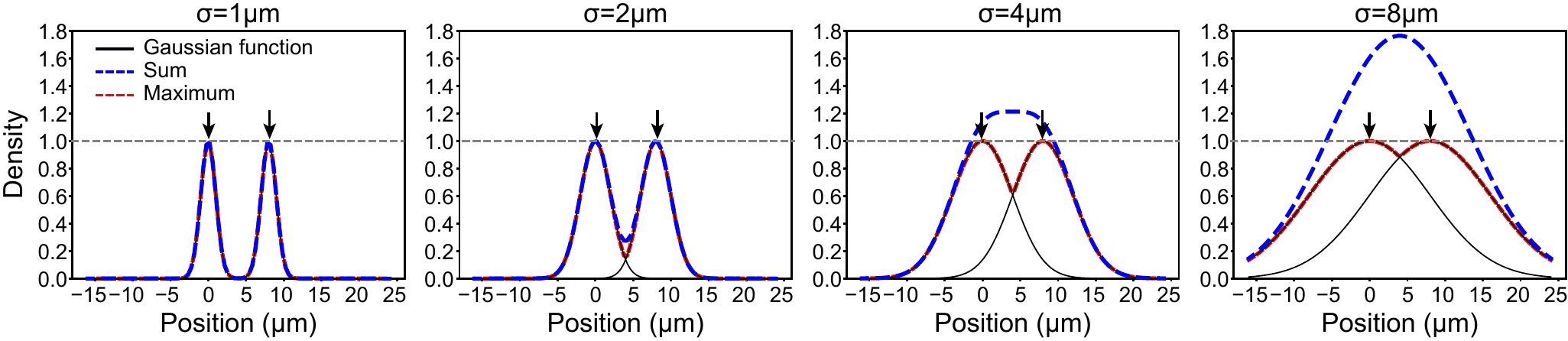}
    \caption{ 
    1D illustration of the compounding of Gaussian kernels with different kernel sizes $\sigma$. 
    Black arrows illustrate the coordinates of two different cells which are one diameter (\SI{8}{\micro\meter}) apart. 
    Compounding of kernels by their maximum (\emph{K}$_{\!\mathrm{max}}$) allows to discern their respective peaks regardless of the $\sigma$ value and the resulting density value is bounded (dashed gray line) within an interval, allowing for unified interpretation of a peak and a fixed dynamic range of GT DMs desirable in CNN training. 
    Meanwhile, compounding by their sum (\emph{K}$_{\!\mathrm{sum}}$) merges peaks into a single one for values of $\sigma$ greater than \SI{2}{\micro\meter}, and leads to density values higher than those of the original Gaussians, hence confounding peak density with amplitude while hampering peak detection and threshold-based methods. 
    }
    \label{suppfig:gaussian_aggregation}
\end{suppfigure}

\begin{suppfigure}[hbt!]
    \centering
    \includegraphics[width=0.7\textwidth]{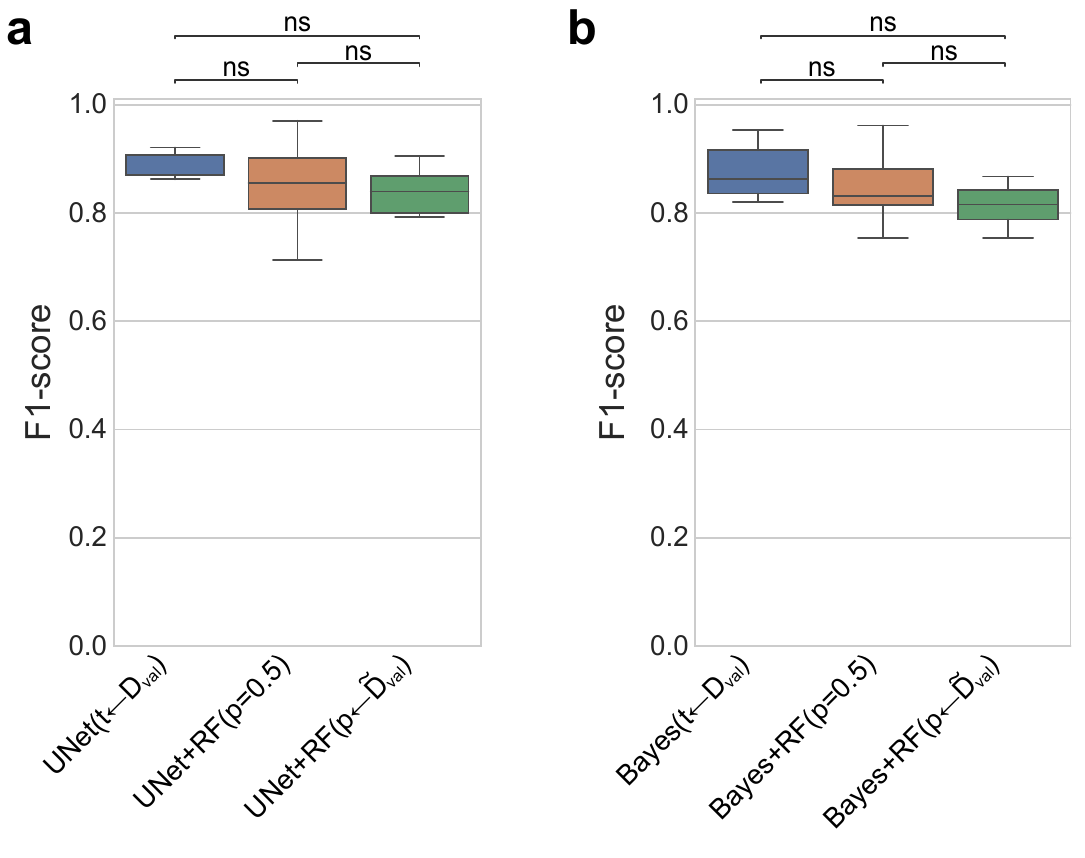}
    \caption{ 
    Comparison of probabilistic models with \emph{RF} using different criteria for for thresholding the probability $p$.
    Under the probabilistic interpretation, $p$ can be set to 0.5, which is named RF(p=0.5).
    For comparison with threshold-based detection methods (UNet or Bayes) where this threshold $t$ is selected from the validation set ($t \leftarrow D_\textrm{val}$), we also consider a RF where the threshold $t$ is selected from its corresponding validation set as $RF(p \leftarrow \tilde{D}_\textrm{val})$.
    These results confirm that detection results are not significantly different with either of these methods, although the probabilistic interpretation of $p=0.5$ produces slightly superior results.
    }
    \label{suppfig:validation_prob}
\end{suppfigure}

\begin{supptable}[hbt!]
    \centering
    \caption{
    Summary of labeled dataset employed for the evaluation of detection methods. 
    }
    \begin{tabular}{c|cc}
    Sample ID & \# patches & \# annotated coordinates \\
    \hline
    1 & 16 & 440 \\
    2 & 24 & 549 \\
    3 & 12 & 171 \\
    4 & 12 & 332 \\
    5 & 16 & 338 \\
    6 & 16 & 566 \\
    7 & 20 & 642 \\
    \hline
    TOTAL & 116 & 3038 \\
    \end{tabular}
    \label{supptab:dataset}
\end{supptable}

\begin{supptable}[hbt!]
    \centering
    \caption{
    Tiling parameters employed for the two strategies presented: \emph{M}$_{\!\mathrm{conv}}$ and \emph{M}$_{\!\mathrm{peak}}$. 
    Sizes are in voxels. 
    }
    \resizebox{\textwidth}{!}{
    \begin{tabular}{l|ccccc}
    \textbf{Tiling strategy} & \textbf{Input ($l_\textrm{in}$)} & \textbf{CNN output ($l_\textrm{out}$)} & \textbf{Output ($l_\textrm{out\_tile}$)} & \textbf{Padding ($l_\textrm{pad}$)} & \textbf{Overlap ($l_\textrm{overlap}$)} \\
    \hline
    \emph{M}$_{\!\mathrm{conv}}$ & $64\times156\times156$ & $24\times116\times116$ & $24\times116\times116$ & $20\times20\times20$ & $40\times40\times40$ \\
    \emph{M}$_{\!\mathrm{peak}}$ & $64\times156\times156$ & $24\times116\times116$ &  $16\times108\times108$ & $24\times24\times24$ & $48\times48\times48$ \\
    \end{tabular}}
    \label{supptab:tiling}
\end{supptable}

\end{document}